%% file: latex12.tex
\documentclass[a4paper, times, 10pt,twocolumn]{article}
\usepackage{latex12}
\usepackage{times}
\usepackage{balance}
\usepackage{rotating}
\usepackage{float}
\usepackage{scalefnt}
\usepackage{multicol,lipsum}
\usepackage{soul}
\usepackage{xcolor}
\usepackage{amsmath,amssymb,amsfonts}
\usepackage{algorithmic}
\usepackage{graphicx}
\usepackage{textcomp}
\usepackage{amsmath}
\usepackage{amsfonts}
\usepackage{amssymb}
\usepackage{calrsfs}
\usepackage[ruled,vlined,linesnumbered]{algorithm2e}
\usepackage{authblk}
\newcommand*{\affaddr}[1]{#1} 
\newcommand*{\affmark}[1][*]{\textsuperscript{#1}}

\begin{document}

\title{Classifier Pool Generation based on a Two-level Diversity Approach}

\author{
Marcos Monteiro\affmark[1], Alceu S. Britto Jr\affmark[1,2], Jean P. Barddal\affmark[1], Luiz S. Oliveira\affmark[3], and Robert Sabourin\affmark[4]\\
\affaddr{\affmark[1]PPGIa, Pontif\'{i}cia Universidade Cat\'{o}lica do Paran\'{a} (PUCPR), Curitiba, PR, Brazil}\\
\affaddr{\affmark[2]State University of Ponta Grossa (UEPG), Ponta Grossa, PR, Brazil}\\
\affaddr{\affmark[3]Federal University of Paran\'a (UFPR), Curitiba, PR, Brazil}\\
\affaddr{\affmark[4]\'Ecole de Technologie Sup\'erieure (\'ETS), Universit\'e du Qu\'ebec, Montr\'eal, QC, Canada}\\
\affaddr{\{marcos, alceu, jean.barddal\}@ppgia.pucpr.br},
\affaddr{lesoliveira@dinf.ufpr.br},
\affaddr{robert.sabourin@etsmtl.ca}

}

\maketitle
\thispagestyle{empty}

\begin{abstract}
This paper describes a classifier pool generation method guided by the diversity estimated on the data complexity and classifier decisions. First, the behavior of complexity measures is assessed by considering several subsamples of the dataset. The complexity measures with high variability across the subsamples are selected for posterior pool adaptation, where an evolutionary algorithm optimizes diversity in both complexity and decision spaces. A robust experimental protocol with 28 datasets and 20 replications is used to evaluate the proposed method. Results show significant accuracy improvements in 69.4\% of the experiments when Dynamic Classifier Selection and Dynamic Ensemble Selection methods are applied.
\end{abstract}

\section{Introduction}
Multiple Classifier Systems (MCS) is an effective way to improve accuracy on many classification problems. 
An MCS is often layered in three steps: (a) pool generation, (b) classifier selection, and (c) integration \cite{Britto2014}.
During \emph{pool generation}, a pool of classifiers is created.
Next, the \emph{selection} step optionally selects a subset of classifiers from the entire pool.
Finally, the \emph{integration} step combines the predictions obtained from the selected classifiers.
The key to tailoring successful ensembles is diversity, as each learner should cast a different prediction for the same input instance, while all learners should be consistent and perform better than a random guesser.
Even though there is no clear correlation between higher diversity and accuracy \cite{kuncheva2002theoretical}, the need for diversity can be explained by assuming a group of experts and their opinions. If the opinions of all experts match, a single expert suffices.

During the classifier selection step, an MCS can be categorized according to the number of classifiers selected.
If a single learner is chosen, the system is referred to as a Dynamic Classifier Selection (DCS) system, whereas a Dynamic Ensemble Selection (DES) potentially picks more than a single classifier \cite{cruz2018dynamic}.
In both DCS and DES, the selection from a pool of classifiers is performed during the test phase, i.e., a specific selection is performed per test sample. 
The rationale behind the dynamic selection is to select the `most appropriate' classifiers to predict each test instance.

In this paper, we propose a novel method for creating a pool of classifiers. 
To reach this goal, we explore two kinds of diversity: (i) in the complexity space, and (ii) in the decision space. The first is related to training classifiers on datasets representing sub-problems with different levels of complexity, while the second is related to generating classifiers that make different errors.
We hypothesize that both diversities can guide pool generation as an optimization problem, which is solved using a multi-objective genetic algorithm. 
The contribution of this work is twofold. 
First, a new pool generation method that trains classifiers on data subsets with different complexity levels is proposed. 
Second, we show the positive impact of the proposed scheme when combined with dynamic selection methods, as the pool covers a wider spectrum of the complexity space, thus making the selection process more efficient.

This work is organized as follows.
Sec. \ref{sec:relatedworks} presents the state-of-the-art ensemble methods. 
Sec. \ref{sec:datacomplexity} presents the data complexity measures since they are at the core of our proposal. 
Sec. \ref{sec:proposal} describes the proposed pool generation method. 
Sec. \ref{sec:analysis} presents the evaluation of the proposed method, and the results are analyzed.
Finally, Sec. \ref{sec:conclusion} concludes this work and delineates ideas for future works.

\section{Related Works} \label{sec:relatedworks}
Even though MCS is widely used nowadays, its foundations go back from seminal papers such as \cite{kuncheva2002theoretical, opitz1999popular}, where the main advantages in terms of prediction rates of this kind of systems are presented and compared to approaches where a single classifier is used.

Seminal approaches for pool generation include Bootstrap Aggregating (Bagging) \cite{breiman1996bagging}, AdaBoost \cite{Freund1996} and Random Subspaces \cite{Ho1998}. Bagging uses the random distribution of problem examples to create new training subsets via resampling. In contrast to Bagging, vertical data partitioning approaches such as Random Subspaces induce diversity in the classification pool as each base learner is trained using a different subset of features, which is drawn from a uniform distribution. Now, in contrast to Bagging and Random Subspaces, Boosting approaches construct learners sequentially, where the construction of a base learner exploits the mislabeled instances observed in the previous rounds.

Souza et al. \cite{souza2017characterization} present another approach for creating a pool of classifiers based on the Oracle upper bound initially described in \cite{kuncheva2002theoretical}.
An Oracle is an optimal model where if any of the base learners correctly classify a test instance, then it is assumed that the final prediction of the MCS is also correct. 
Even though the authors observed 100\% accuracy rates in several datasets, they showed that no existing dynamic classifier selection techniques achieves this upper bound.

Related to our approach, different studies \cite{Ho2002, luengo2015} have shown how complexity metrics improve the creation and validation of MCS'.
For instance, authors in \cite{luengo2015} use complexity metrics to identify the competence region of heterogeneous classifiers. 
They concluded that complexity metrics improve dynamic classifier selection upon competence regions.
Complexity measures were also combined with meta-features in \cite{Cruz2015} to create pools of classifiers in the META-DES framework.
Finally, Brun et al. \cite{Brun2016} describe how specific metrics can be used to create pools of classifiers and guide the dynamic classifier selection process.

\section{Data Complexity Measures} \label{sec:datacomplexity}
Diversity and data complexity measures are the core of the proposed method. Due to space constraints, in this paper we only describe the main data complexity measures. Besides, an important contribution on diversity measures can be found in \cite{kuncheva2003}.

According to \cite{HO2006}, complexity measures analyze the difficulty of a classification problem. The complexity of a problem is due to 3 factors: (i) class ambiguity, (ii) data sparsity, and (iii) difficulty of separating instances in their respective classes. 
Class ambiguity regards the poor distinction between classes given the features available. 
Conversely, datasets with a small ratio between instances and features or class imbalance form hard problems for the classifier to learn from.

To quantify how hard a classification problem is, authors in \cite{Lorena2018} proposed a taxonomy to group complexity metrics in the following families: Overlapping, Linearity, Neighborhood, Network, Dimensionality, and Class Balance. 
In this work, we focus on the Overlapping and Neighborhood families as their computational cost is smaller and yielded interesting results in previous studies in MCS' \cite{Brun2016}.

\subsection{Overlapping measures} 

This group of measures analyzes the feature space of a problem to check if classes overlap, i.e., how `close' the distribution of attributes for instances from different classes are to each other. The measures belonging to this group are used when the features available are numeric and include the following metrics: F1, F1v, F2, F3, and F4. 
\noindent\textbf{Maximum Fisher’s Discriminant Ratio (F1)} quantifies the distance between the centroids of two classes. This metric is based on the mean and standard deviation of each attribute \cite{Orriols-Puig2010}. 
\noindent\textbf{The Directional-vector Maximum Fisher’s Discriminant Ratio (F1v)} \cite{malina} adapts F1 to compute the class overlap in multi-class problems. First, the dataset is orthogonally projected in a feature space $\vec{d}$, in which instances are maximally discriminated.
This process is achieved via matrix decomposition and eigenvectors and eigenvalues computation.
\noindent\textbf{Volume of Overlap Region (F2)}~\cite{Lorena2018} assumes that class overlap is given by the overlap region observed between classes in a single attribute.
Their metric, hereafter referred to as F2, is computed based on the minimum and maximum value observed per attribute and per class.
The overlap interval is calculated by normalizing the range of values in both classes.
\noindent\textbf{Feature Efficiency (F3)} represents the dimensionality of the attribute distribution.
According to \cite{Orriols-Puig2010}, the result of F3 is obtained by following the heuristic: for each of the instances of each one of the classes, it is evaluated if it contains any attribute that has value in a region belonging to another class, which can cause it to be misclassified. 
If this condition holds, the instance is marked. 
F3 is the ratio between the number of instances marked and the total number of instances. \textbf{Collective Feature Efficiency (F4)} selects the attribute that best separates the classes. The instances correctly separated with this attribute and the attribute itself are removed from the dataset, and then the process is repeated. 
This loop is repeated until all the instances can be labeled or until all the attributes have been analyzed. 
F4 corresponds to the proportion of instances that can be discriminated \cite{Orriols-Puig2010, Lorena2018}.
\subsection{Neighborhood measures}
Neighborhood measures account for the distance between examples and/or attributes to characterize borderline regions between classes and their separability. 
\noindent\textbf{Fraction of Borderline Points (N1)} is the percentage of data samples that are close to the boundary observed by samples of different classes.
To find these limits, the N1 metric uses a Minimum Spanning Tree, which generates a tree connecting examples of the problem.
Each connection between two samples from distinct classes represents the boundary, which is used to compute the number of samples in the decision boundary \cite{Ho2002, Lorena2018}. 
\noindent\textbf{Ratio of Intra/Extra Class Nearest Neighbor Distance (N2)} is given by the ratio between intra and extra class nearest neighbors \cite{Ho2002, Lorena2018}.
\noindent\textbf{Error Rate of the Nearest Neighbor Classifier (N3)} refers to the error rate of the 1NN classifier estimated using leave-one-out \cite{Lorena2018}.
\noindent\textbf{Non-Linearity of the Nearest Neighbor Classifier (N4)} uses the training dataset to synthesize novel instances for the nearest neighbor assessment. This new dataset is created via the interpolation of examples within each class. The error rate of the KNN classifier on top of this newly generated dataset determines N4 \cite{Ho2002, Orriols-Puig2010}. 
\textbf{Fraction of Hyper-spheres Covering Data (T1)} is calculated after the creation of hyper-spheres that are centered in a randomly picked instance of a class. These hyper-spheres are increased until they touch instances from other classes. T1 is the ratio between the number of hyper-spheres required to cover a class and the total number of instances in the dataset \cite{Lorena2018, Ho2002}.

\section{Proposed Method}\label{sec:proposal}
\vspace{-0.5em}
This section introduces our proposal called \textit{Classifier Pool Generation based on Diversity in the Decision and Complexity Spaces} (PGDCS). 
PGDCS has two steps: (i) selection of data complexity metrics, followed by (ii) pool generation based on Multi-objective Genetic Algorithm (MOGA).
In the first step, given a dataset, PGDCS evaluates the behavior of different data complexity metrics and determines which should be used according to their variability.  
As described in Sec. \ref{sec:datacomplexity}, we used two families of complexity metrics and a metric from each group is selected to avoid the selection of correlated metrics.
During the second step, a multi-objective genetic algorithm is used to generate a pool of classifiers given the data complexity metrics selected.
The objective is to create a pool of classifiers trained on data subsets with high diversity in terms of levels of difficulty and making different errors (decision diversity).
Each step is described below.

\subsection{Selection of data complexity metrics}
First, the dataset is divided into training, validation, and testing sets.
At each iteration of the proposed algorithm, the $N$ subsets are randomly selected from the replacement training dataset.
For each subset, all the complexity measures described previously in Section \ref{sec:datacomplexity} are computed.
Then, each metric dispersion (standard deviation) among all subsets of data is calculated in a paired way.
The metric that shows the highest dispersion receives a vote in each iteration of the proposed algorithm.
Finally, our method selects the measures with the most votes from the Overlapping ($cm_1$) and Neighborhood ($cm_2$) families.
The logic behind the selection of the most dispersed metric is to have more flexibility in the second stage when we reach sub-problems with different levels of difficulty.
The metric with the most votes in each family is selected for the classification problem analyzed.
The metrics selected in this step are used as part of the optimized objective function during the next step of the proposed method.

\subsection{Pool generation based on MOGA}
The second step regards the generation of the classification pool using NSGA2 \cite{Deb2002}. Given the training dataset, different sub-problems are created by re-sampling the training instances. 
In the proposed optimization process, each subset of samples is a chromosome and each gene is an instance.

The crossover operator generates a new individual (data subset) by exchanging instances from subsets $S_i$ and $S_j$.
Alg. 1 presents this process. 
In lines 2 and 3, two arbitrary data subsets $S_i$ and $S_j$ are randomly chosen within the population $P$ and the resulting individual $S_{out}$ is initialized as an empty structure (line 4).
Next, the variables \texttt{start} and \texttt{end} are randomly set according to an uniform distribution $\mathcal{U}$ (lines 5 - 6) and are used to restrict the interval in which instances are drawn from $S_i$ or $S_j$ (lines 7-11).

\begin{algorithm}[!t]
\caption{Crossover}
  \DontPrintSemicolon
  \SetAlgoLined
  \SetKwInOut{Input}{Input}\SetKwInOut{Output}{Output}
\Input{$P$, $\mathcal{U}$}
  \Output{$S_{out}$}
  \SetKwFunction{FMain}{Crossover }
  \SetKwProg{Fn}{Function}{:}{\KwRet{$S_{out}$}}
  \Fn{\FMain{$P, \mathcal{U}$}}{
           $S_i \leftarrow $ randomly select an individual from $P$\\
           $S_j \leftarrow$ randomly select an individual from $P \setminus \{S_1\}$\\
           $S_{out} \leftarrow \emptyset$\\
           \textit{\texttt{start}} $\leftarrow \mathcal{U}(0, x_n)$\\
           \textit{\texttt{end}} $\leftarrow \mathcal{U}(\text{start}, x_n)$\\
          
          \For {$k$ \textbf{to} $\max\left(\vert S_1 \vert, \vert S_2 \vert\right)$ }{
             \uIf {$k \leq$ \texttt{start} \textbf{or} $i \geq$ \texttt{end}}{
                 $S_{out}[k] \leftarrow S_{i}[k]$}
             \lElse{
                 $S_{out}[k] \leftarrow S_{j}[k]$}
           }
  }
  \label{alg:Crossover}
\end{algorithm}

During mutation, a data subset (chromosome) exchanges a gene, i.e. an instance of the data subset. 
To this end, a ``donor'' and a ``receiving'' subsets are randomly picked, and a gene from the receiving subset is replaced by an instance from the donor. 

The optimization process occurs based on three objectives. 
The first two explore diversity in the complexity space by using the metrics selected in the first phase of the method. 
The idea is to maximize the average distance (dispersion) among the data subsets considering each complexity metric computed in a pairwise manner. 
The third objective is to maximize the diversity in the decision space by using the Double Fault ($DF$) metric computed on the decisions of the classifiers trained on each data subset also in a pairwise manner.
According to \cite{ruta2005classifier}, superior results in majority vote ensembles were obtained via $DF$ optimization. 

In preliminary tests, we used only one complexity measure, yet, the pool lost diversity.
This is due the fact that certain instances have much higher or lower complexity values, and thus, these were selected multiple times in several bags as their occurrence increased the overall fitness values. 
Therefore, higher complexity values were achieved, yet, with the drawback of decreasing the pool diversity.
Therefore, our rationale is to apply different complexity metrics combined with a diversity metric and make all of them equally weighted in the genetic algorithm optimization process.

Three objectives are computed considering per $S_i$:
- \textbf{$\Phi_{cm_1}[S_i]$} and \textbf{$\Phi_{cm_2}[S_i]$}: the dispersion values of $S_i$ considering the complexity measures $cm_1$ and $cm_2$. To obtain the dispersion, the algorithm computes $\varphi_{cm_j}[S_i]$, which is the value of the metric $cm_j$ for each $S_i$. Next, in a pairwise manner, the difference between $S_i$ and all other data subset complexity values $\varphi_{cm_j}[S_k]$ is computed as denoted in Eq. \ref{eq:disp}, where $N$ is the total number of subsets. The dispersion $\Phi_{cm_j}[S_i]$ stores the average distance of the subset $S_i$ with respect to all other subsets in the complexity space regarding $cm_j$. The idea is to maximize these dispersion values to better cover the problem complexity space.
\vspace{-0.16cm}
\begin{equation}
    \Phi_{cm_j}[S_i] = \frac {\sum_ {k = 1} ^ N abs(\varphi_{cm_j}[S_i] - \varphi_{cm_j}[S_k])} {N-1}
\label{eq:disp}
\end{equation}

\noindent- \textbf{$DDV[C_i]$} is the Decision-based Diversity Value of a classifier $C_i$ computed using the Double Fault diversity measure. The idea is to maximize the diversity w.r.t. classifiers decisions. First, a classifier ($C_i$) is trained for each subset of data ($S_i$) representing a subproblem with a given complexity level. Next, the algorithm calculates $DF (C_i, C_j)$, which is the Double Fault diversity measure per classifier using the validation set. Finally, a final diversity value is estimated for $S_i$ in a pairwise manner as denoted in Eq. \ref{eq:parwaisedf}, where $N$ is the number of classifiers.
\vspace{-0.16cm}    
\begin{equation}
    DDV[C_i] = \frac{\sum_{j = 1}^{N}DF(C_i, C_j)}{N-1}
\label{eq:parwaisedf}
\end{equation}

Alg. 2 describes the fitness function of our method, in which the inputs are the population $P$ and the corresponding number of individuals ($\mu$). 
In the first loop (lines 2 to 6), the complexity values related to the selected metrics of overlapping and neighborhood of each subset $S_i$ are computed and stored in $\varphi_{cm_1}[S_i]$ and $\varphi_{cm_2}[S_i]$, respectively. 
Besides, a classifier is trained for each $S_i$ considering a previously defined base classifier.
In the second loop (lines 7 to 11), in a pairwise manner, the algorithm computes for each $S_i$ the dispersion $\Phi_{cm_1}$ and $\Phi_{cm_2}$ related to the measures $cm_1$ and $cm_2$ using Eq. \ref{eq:disp}. 
In addition, the Double Fault for each $S_i$ is computed using Eq. \ref{eq:parwaisedf}. 
Finally, the global dispersion $G_{disp}$ is evaluated using Eq. \ref{eq:G}, which represents the general dispersion of the whole population.

The stop criterion of the proposed optimization process is the number of generations. 
However, for each generation we keep a global descriptor $F$ which is composed of the three feature values: $\Phi_{cm_1}$, $\Phi_{cm_2}$, and $DDV[C_i]$. 
The best solution is defined considering the generation that has a higher dispersion considering the $F$ descriptor computed as denoted in Eq. \ref{eq:G}, where $G_{\textrm{disp}}(t)$ is the global dispersion on generation $t$, $N$ is the number of bags/classifiers and $F_n$ is the number of objective functions being assessed. 

During the MOGA execution, certain complexity metrics can reach their dispersion limit, i.e., their variability is zero. 
MOGA overlooks the limit of a specific metric as the remainder still have room for dispersion. 
Therefore, we use Eq. \ref{eq:G} to select the generation in which the three objectives are globally more dispersed.

\vspace{-0.10cm}  
\begin{equation}
     G_{\textrm{disp}}(t) = \frac {\sum_ {j = 1} ^ {N} \sqrt {\sum_ {h = 1} ^ {F_n} (F[i, h] - F[j, h]) ^ 2}} {N-1}
     \label{eq:G}
\end{equation}

\begin{algorithm}[!t]
\caption{Fitness}
  \DontPrintSemicolon
  \SetAlgoLined
  \SetKwInOut{Input}{Input}\SetKwInOut{Output}{Output}
  \Input{$P$, $\mu$}
  \Output{$F$}
  \SetKwFunction{FMain}{Evaluate}
  \SetKwProg{Fn}{Function}{:}{\KwRet{$F$}}
  \Fn{\FMain{$P$, $\mu$}}{
           \For {each $S_i$ \textbf{in} $P$}{
                compute metric $cm_1$ as $\varphi_{cm_1}[S_i]$\\
                compute metric $cm_2$ as $\varphi_{cm_2}[S_i]$\\
                train a classifier $C_i$ on $S_i$\\
        }
        \For {i \textbf{in} [1..N]}{
                compute $F[i][0]$ as $\Phi_{cm_1}[S_i]$ (Eq. \ref{eq:disp}) \\
                compute $F[i][1]$ as $\Phi_{cm_2}[S_i]$ (Eq. \ref{eq:disp}) \\
                compute $F[i][2]$ as $DDV[C_i]$ (Eq. \ref{eq:parwaisedf}) 
        }
  }
  \label{alg:fitness}
\end{algorithm}

Alg. 3 presents the multi-objective genetic algorithm used, which receives as input: the number of generations $\psi$, the population size $\mu$, the number of new bags after crossover and mutation $\theta$, and the offspring size $\gamma$. 
At $t=0$ (line 2), the first population of data subsets $P(t)$ is created with size $\mu$ in line 3. 
In line 4, $F(t)$ receives the 3 fitness values of each individual in the generation $t$, which are computed using Alg. 2. 
The loop in lines 5-18 depicts the number of iterations of the genetic algorithm.
Line 6 depicts the crossover and mutation steps, which are followed by the offspring definition in Line 8.
In line 9, the global dispersion $\upsilon$ is computed taking account for the three fitness values.
Lines 10 to 16 store the best results across the generations based on the global dispersion value. 
Our hypothesis is that the generation with the highest global dispersion $\varrho$ better covers the problem complexity space and a pool trained on such data has higher diversity in the decision space.

\begin{algorithm}[!t]
    \caption{Genetic Algorithm}
  \DontPrintSemicolon
  \SetAlgoLined
  \SetKwInOut{Input}{Input}\SetKwInOut{Output}{Output }
  \Input{$\psi $, $\mu$, $\theta$, $\gamma$}
  \Output{$\varrho$}
  \SetKwFunction{FMain}{GA}
  \SetKwProg{Fn}{Function}{:}{\KwRet{$\varrho$}}
  \Fn{\FMain{$\psi $, $\mu$, $\theta$, $\gamma$}}{
        $t \gets 0$\;
        $P(t) \gets$ initialize($\mu$)\;
        $F(t) \gets$ evaluate($P(t),\mu$)\;
        \While{$t \neq \psi $}{
            $P'(t) \gets$ crossover and mutation($P(t), \theta$)\;
            $F(t)\gets$ evaluate($P'(t), \gamma$)\;          
            $P(t+1) \gets$ select($P'(t), P(t), F(t), \mu$)\;
            $\upsilon \gets$ $G_{disp}(t)$  (Eq. \ref{eq:G})
            \BlankLine
            \uIf{$t=0$}{
               $\varrho \gets P(t+1)$\\
               $temp \gets \upsilon$
            }
            \ElseIf{$temp < \upsilon$}{
                 $\varrho \gets P(t+1)$\\
                 $temp \gets \upsilon$
                }
            
            $t\gets$ $t+1$\; 
        
        } 
  }
  \label{alg:GA}
\end{algorithm}

\begin{figure}[!t]
    \centering
    \resizebox{\columnwidth}{!}{
        \includegraphics[scale=0.6]{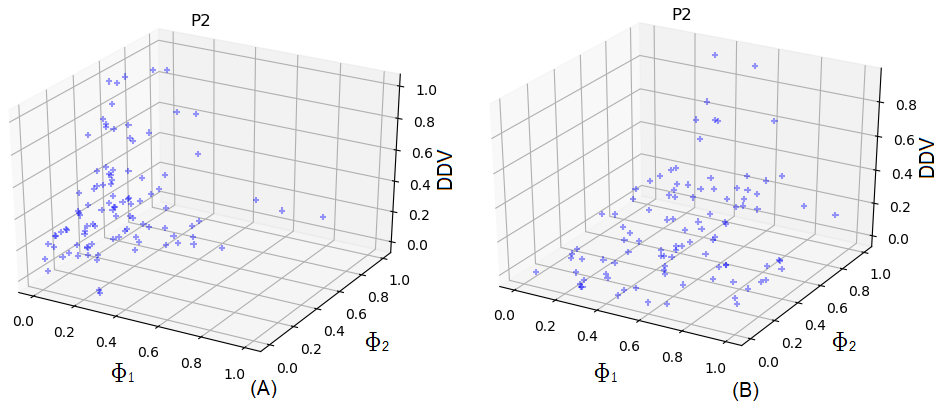}
    }
    \caption{Dispersion in the complexity space for P2 dataset: (A) first and (B) best generation of the PGDCS.}
    \label{fig:DistanceP23d}
    \vspace{-1.5em}
\end{figure}


\subsection{A note on complexity analysis}
Assuming a dataset with $n$ instances, $f$ features, and $y$ classes, the first part of proposed method is run $r$ times.
In each run, a number of bags $N$ is created with random sampling of the original dataset.
Next, the complexity metrics are computed along with their standard deviation, and their aggregated complexity is $O(f^{3}y^{2})$ (proof in \cite{Jain2014}).
Consequently, the complexity of the first step is $O(rn(f^{3}y^2 + N))$.
The second step regards the genetic algorithm optimization process, in which NSGA2 \cite{Deb2002} was applied.
This process is repeated $r^{\prime}$ times, such that each run encompasses $g$ generations.
For each generation $g$, NSGA2 has a time complexity of $O(\mu\log\mu^{F_{n}-2})$, where $F_n$ is the number of objectives being optimized.
Nonetheless, the computation of the objectives depend on complexity metrics, which as depicted above, have an $O(f^{3}y^{2})$ cost.
Therefore, the time complexity of the second step of the proposed method is of $O(g \times r^{\prime} \times (\mu\log\mu^{F_{n}-2} f^{3}y^2))$.
As a result, the overall time complexity of the proposed method is $O(rn(f^{3}y^2 + N) + g \times r^{\prime} \times (\mu\log\mu^{F_{n}-2} f^{3}y^2))$.


\section{Analysis of the PGDCS Method} \label{sec:analysis}

To evaluate the PGDCS method, the pool members were combined using the majority voting rule (MVR) and using 6 well-known methods of dynamic selection: LCA, OLA and Rank \cite{woods1997} as Dynamic Classifier Selection (DCS), and KNORA-U, KNORA-E \cite{AlbertKo} and META-DES \cite{Cruz2015} as Dynamic Ensemble Selection (DES).
All of these techniques are publicly available as part of the \texttt{deslib} library \cite{JMLR}. 
Furthermore, in the following results, we used a linear Perceptron as base classifier, which in spite of its simplicity, has achieved superior results when dynamic selection is used \cite{cruz2015deep}. 

\input{problem_table.tex}

Tab. \ref{tb:problem} details the datasets used. 
Each experiment was split into training, validation, and testing sets using 50\%, 25\%, and 25\% of the dataset, respectively. 
The pool size used in our experiments was $N=100$ and 20 executions of each experiment were performed.

\subsection{Selection of data complexity metrics}
\vspace{-0.14cm} 
In this step, PGDCS selects metrics with higher variability in the overlapping and neighborhood families.
To avoid the impact of randomness, this process was repeated 20 times.
Tab. \ref{tab:vote} summarizes the voting results of each metric per experiment.
Across datasets, no metric dominates others, thus showing that this definition is dataset dependent.

\input{vote_table.tex}

\subsection{Pool generation based on MOGA}
After selecting the most promising complexity measures per dataset, the data bags are generated using multi-objective genetic algorithms using 20 generations.
In the first generation of the optimization process, the method generates $\mu$ data subsets (initial population) and each of these contain 50\% of the instances available selected randomly. The number of generations ($\psi$) is 20, the population size ($\mu$) is 100, the number of children ($\theta$) is 100; while the offspring size ($\gamma$) is 100.



A detailed analysis regarding the pool generation process was done using the P2 dataset. 
The result can be seen in Fig. \ref{fig:DistanceP23d}. 
Each element in Fig. \ref{fig:DistanceP23d} represents a subset: the x-axis ($\Phi_1$) demonstrates the average pairwise distance of each measure of complexity, as well as the z-axis ($\Phi_2$), while the y-axis ($DDV$) represents the diversity (Double Fault) of each data subset. 
We see in Fig. \ref{fig:DistanceP23d} (A) the individuals (subsets of data) in the first generation, while in Fig. \ref{fig:DistanceP23d} (B) we have an overview of the individuals in the best generation. 
The difference between them is clear, as in (B) the bags are spread compared to what is observed in (A), thus showing that the genetic algorithm improves diversity.


The average accuracy and the standard deviation of 20 replications for each classification problem and a comparison with the Bagging \cite{breiman1996bagging} method is presented in Tab. \ref{tab:tabresult}. 
We applied the Wilcoxon paired test \cite{wilcoxon1945individual} to verify the existence of a statistically significant difference between the pool of classifiers generated by PGDCS and the pool of classifiers generated using the Bagging method per dataset. 
The results marked with an asterisk do not show statistic differences for a 95\% confidence level, whereas the bold values are highest, on average, per experiment and selection method.

\input{result_table_max_dist.tex}

\begin{figure}[!t]
    \centering
    \resizebox{\columnwidth}{!}{
    \includegraphics{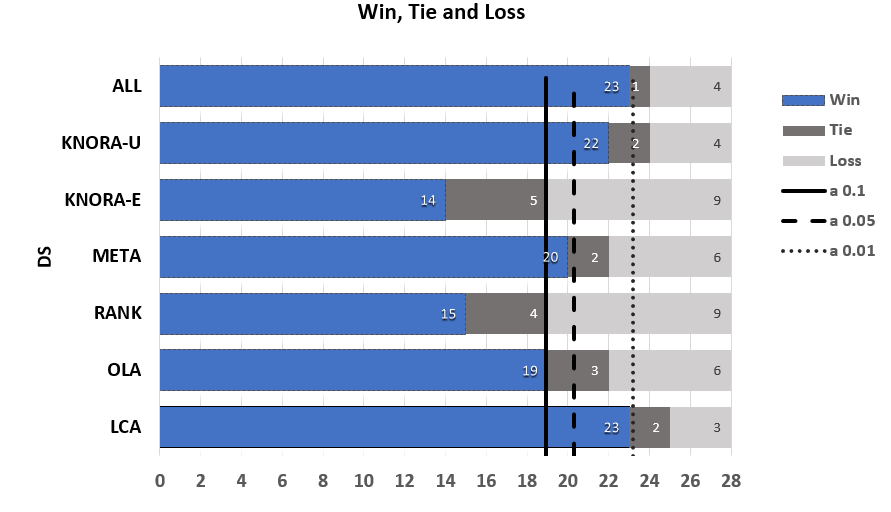}
    }
    \caption{Pairwise comparison Bagging and proposed method using DCS, DES, and MVR across 20 replications}  
    \label{fig:wtl}
    \vspace{-2.0em}
\end{figure}

Overall, we performed 196 experiments, our method won at 69.4\%, tied at 9.6\% and lost at 20.9\%. 
Fig. \ref{fig:wtl} presents a win-tie-loss analysis of the results. 
The critical level ($Nc$) is represented in the Fig. \ref{fig:wtl} with levels of significance of $\alpha=\{0.1, 0.05, 0.01\}$. 
The number of wins plus half the number of draws must be greater than $Nc$ (Eq. \ref{eq:nc}), where $N_{exp}=28$ (number of experiments), $Nc=$\{18.9, 20.3, 23.2\} for the significant levels $\alpha=$\{0.1, 0.05, 0.01\} respectively.

\vspace{-0.2cm} 
\begin{equation}
      Nc=\frac{N_{exp}}{2} + Z_\alpha  \frac{\sqrt{N_{exp}}}{2}
      \label{eq:nc}
\end{equation}

We see in Fig. \ref{fig:wtl}, when compared to the pool generated with Bagging both using majority voting rule (named as ALL classifiers), the proposed method won, tied and lost in 82.1\%, 3.5\% and 14.2\% of the experiments, respectively. For $\alpha$ 0.1 and 0.05, our method was lower than $ N_c $, in the Rank and Knora-E selection methods. In the other selection and combination methods, we exceed the critical level for $\alpha$ 0.1 and 0.05. Very interesting results were also observed in the experiments using dynamic selection methods. We evaluated the impact of the generated pool on six dynamic schemes for selection of classifiers, being three Dynamic Classifier Selection (DCS) and three Dynamic Ensemble Selection (DES) methods. 

When considering the DCS methods, the proposed method won, tied and lost in 67.8\%, 10.7\% and 21.4\% of the experiments, respectively. 
Similar results were observed w.r.t. DES, where the proposed method won, tied and lost in 66.6\%, 7.2\%, and 26.2\%, respectively. 

Regarding the scenarios in which our proposal underperforms Bagging, we note that MOGA reaches its dispersion limit.
Therefore, the bags created do not significantly differ in terms of complexity and the overall ensemble accuracy is jeopardized.
As a result, we plan to account for accuracy rates during the MOGA's evolution process in future works.



\section{Conclusion} \label{sec:conclusion}

In this paper, we proposed a new approach for creating a diverse pool of classifiers. 
PGDCS uses diversity in both complexity and decision spaces to guide the classifier pool generation process.
A robust experimental protocol with 28 datasets and 20 runs confirmed our hypothesis that data complexity and diversity in terms of classifier decisions can leverage pool generation improves existing strategies.
As a result, we observed that our proposal outperforms existing approaches in 69.4\% of the cases.
Future works will consider different strategies to select the best generation of the optimization process as well as to add new complexity measures.

\balance

\bibliographystyle{latex12}
\bibliography{references}
\balance

\end{document}

%% file: problem_table.tex
\begin{table}[!t]
    \small
    \scriptsize
    \scalefont{0.8}
    \centering
    \setlength{\tabcolsep}{2.2pt}
    \caption{Datasets used in the experiment}
    \label{tb:problem}
    \begin{tabular}{llcccc}
    \hline 
    \textbf{No.} &\textbf{Datasets} & \textbf{Instances} & \textbf{Features} & \textbf{Classes} & \textbf{Source} \\
    \hline
    01 & Australian &690 & 14 & 2 & UCI \cite{uci}     \\
    02 & Banana & 2000 & 2  & 2   & PRTools \cite{duinprtools} \\
    03 & Blood & 748 &  4  & 2       & UCI \cite{uci}     \\
    04 & CTG & 2126 &  21  & 3       & UCI\cite{uci}     \\
    05 & Diabetes & 766 & 8  & 2       & UCI\cite{uci}     \\
    06 & Faults & 1941 &  27  & 7       & UCI \cite{uci}    \\
    07 & German & 1000 &   24 & 2       & STATLOG \cite{king1995statlog} \\
    08 & Haberman  & 306 & 3   & 2       & UCI  \cite{uci}   \\
    09 & Heart & 270 &  10& 2      & STATLOG \cite{king1995statlog} \\
    10 & ILPD & 583 &  34& 2       & UCI \cite{uci}    \\
    11 & Ionosphere &351 & 16 & 2       & UCI \cite{uci}    \\
    12 & Laryngeal1 & 213& 16 & 2       & LKC  \cite{kuncheva2004ludmila}   \\
    13 & Laryngeal3 & 353 & 2 & 3       & LKC \cite{kuncheva2004ludmila}    \\
    14 & Lithuanian & 2000 & 2 & 2       & PRTools \cite{duinprtools} \\
    15 & Liver  & 345 &     6 & 2       & UCI \cite{uci}    \\
    16 & Mammo  & 830 &      5 & 2       & KEEL \cite{alcala2011keel}   \\
    17 & Monk   & 432 &     6 & 2       & KEEL \cite{alcala2011keel}   \\
    18 & Phoneme & 5404 &    5 & 2       & ELENA \cite{jutten2002enhanced}  \\
    19 & P2 & 5000 & 2&2 & \cite{valentini2005experimental}\\
    20 & Segmentation & 2310 & 19&  7       & UCI  \cite{uci}   \\
    21 & Sonar  &208 &     60 & 2       & UCI  \cite{uci}   \\
    22 & Thyroid & 692 &     16 & 2       & LKC \cite{kuncheva2004ludmila}    \\
    23 & Vehicle & 846 &    18 & 4       & STATLOG \cite{king1995statlog} \\
    24 & Vertebral & 300 &  6 & 2       & UCI \cite{uci}    \\
    25 & WBC & 569 &       30 & 2       & UCI \cite{uci}    \\
    26 & WDVG & 5000 &       21 & 3       & UCI \cite{uci}    \\
    27 & Weaning & 302 &   17 & 2       & LKC \cite{kuncheva2004ludmila}    \\
    28 & Wine  & 178&      13 & 3       & UCI \cite{uci}    \\
    \hline \
  \end{tabular}
  \vspace{-0.5cm}
\end{table}


%% file: vote_table.tex
\begin{table}[!t]
\caption{Voting results for measures with highest standard deviation}
\centering
\scriptsize
\scalefont{0.8}
\setlength{\tabcolsep}{2.2pt}
\begin{tabular}{lccccc|ccccccc}
\hline
\textbf{Dataset} & \multicolumn{1}{c}{\textbf{F1}} & \multicolumn{1}{c}{\textbf{F1v}} & \multicolumn{1}{c}{\textbf{F2}} & \multicolumn{1}{c}{\textbf{F3}} & \multicolumn{1}{c|}{\textbf{F4}} & \multicolumn{1}{c}{\textbf{N1}} & \multicolumn{1}{c}{\textbf{N2}} & \multicolumn{1}{c}{\textbf{N3}} & \multicolumn{1}{c}{\textbf{N4}} & \multicolumn{1}{c}{\textbf{T1}} & \multicolumn{1}{c}{\textbf{LSC}} \\ \hline
\textbf{Australian} & \textbf{4} & 3 & 1 & 2 & 0 & 0 & 0 & 2 & \textbf{5} & 1 & 2 \\ 
\textbf{Banana} & 0 & 2 & 3 & \textbf{4} & 1 & 1 & 0 & 3 & 0 & \textbf{4} & 2 \\ 
\textbf{Blood} & 2 & 0 & \textbf{5} & 3 & 0 & \textbf{3} & 1 & 1 & 2 & 1 & 2 \\ 
\textbf{CTG} & \textbf{4} & 3 & 2 & 1 & 0 & 1 & \textbf{3} & 2 & 1 & 2 & 1 \\ 
\textbf{Diabetes} & 1 & 1 & 2 & \textbf{5} & 1 & 1 & 3 & 1 & 0 & 1 & \textbf{4} \\ 
\textbf{Faults} & 0 & \textbf{10} & 0 & 0 & 0 & 1 & 1 & \textbf{4} & 1 & 1 & 2 \\ 
\textbf{German} & 0 & 0 & \textbf{9} & 1 & 0 & 3 & 1 & 0 & \textbf{4} & 1 & 1 \\ 
\textbf{Haberman} & 2 & 2 & \textbf{4} & 0 & 2 & 0 & 1 & 3 & \textbf{4} & 1 & 1 \\ 
\textbf{Heart} & 0 & 0 & 0 & 0 & \textbf{10} & 0 & 0 & 0 & \textbf{6} & 3 & 1 \\ 
\textbf{ILPD} & 1 & 3 & 0 & \textbf{5} & 1 & \textbf{4} & 2 & 2 & 1 & 0 & 1 \\ 
\textbf{Ionosphere} & \textbf{7} & 2 & 0 & 1 & 0 & 0 & 1 & 3 & \textbf{5} & 1 & 0 \\ 
\textbf{Laryngeal1} & 1 & 0 & 0 & 1 & \textbf{8} & 0 & 2 & 2 & \textbf{4} & 0 & 2 \\ 
\textbf{Laryngeal3} & 0 & 3 & 0 & 2 & \textbf{5} & 2 & \textbf{4} & 2 & 1 & 0 & 1 \\ 
\textbf{Lithuanian} & 0 & 0 & 2 & \textbf{6} & 2 & 2 & 2 & 1 & 1 & 1 & \textbf{3} \\ 
\textbf{Liver} & 1 & 5 & 2 & 0 & 2 & 2 & 1 & \textbf{3} & 2 & 2 & 0 \\ 
\textbf{Mammo} & 1 & 2 & \textbf{7} & 0 & 0 & 2 & 1 & 1 & \textbf{3} & 2 & 1 \\ 
\textbf{Monk} & 0 & 0 & 4 & 0 & \textbf{6} & 1 & \textbf{4} & 1 & 2 & 2 & 0 \\ 
\textbf{P2} & 2 & \textbf{5} & 3 & 0 & 0 & 1 & 2 & \textbf{4} & 2 & 0 & 1 \\
\textbf{Phoneme} & 1 & \textbf{4} & 1 & 2 & 2 & 1 & 3 & \textbf{4} & 1 & 0 & 1 \\ 
\textbf{Segmentation} & 0 & \textbf{10} & 0 & 0 & 0 & 0 & 2 & \textbf{4} & 1 & 2 & 1 \\ 
\textbf{Sonar} & \textbf{5} & 2 & 0 & 1 & 2 & 2 & 0 & 2 & \textbf{5} & 1 & 0 \\ 
\textbf{Thyroid} & 2 & 0 & 0 & \textbf{5} & 3 & 0 & 2 & 2 & 1 & 0 & \textbf{5} \\
\textbf{Vehicle} & 0 & 0 & 0 & 1 & \textbf{9} & 1 & 0 & 3 & 0 & \textbf{4} & 2 \\ 
\textbf{Vertebral} & 1 & 2 & 1 & 0 & \textbf{6} & 2 & \textbf{4} & 2 & 2 & 0 & 0 \\ 
\textbf{WBC} & 3 & 2 & 0 & \textbf{5} & 0 & 1 & 2 & 0 & \textbf{3} & 2 & 2 \\ 
\textbf{WDVG} & \textbf{5} & 0 & 0 & 2 & 3 & 3 & 1 & 2 & 0 & \textbf{4} & 0 \\ 
\textbf{Weaning} & 3 & 3 & 0 & \textbf{4} & 0 & 0 & 2 & 0 & \textbf{6} & 0 & 2 \\ 
\textbf{Wine} & 1 & 1 & 0 & \textbf{8} & 0 & 4 & 0 & \textbf{5} & 0 & 1 & 0 \\ 
 \hline
 \textbf{Average} & 1.7 & 2.3 & 1.6 & 2.1 & 2.3 & 1.4 & 1.6 & 2.1 & 2.3 & 1.3 & 1.4 \\ \hline
\end{tabular}
\label{tab:vote}
\vspace{-0.5cm}
\end{table}

%% file: result_table_max_dist.tex
\begin{table*}[!t]
\tiny
\caption{Comparison of PGDCS and Bagging considering the fusion of all classifiers (MVR) and different DCS  and DES methods. "*" stands for no statistical difference}
\label{tab:tabresult}
\resizebox{\textwidth}{!}{
\scalefont{0.6}
\centering
\tiny
\setlength{\tabcolsep}{2.2pt}
\begin{tabular}{ l c c c c c c c c c c c c c c}

\hline
\textbf{} & \multicolumn{ 2}{c}{\textbf{MVR}} & \multicolumn{ 2}{c}{\textbf{LCA}} & \multicolumn{ 2}{c}{\textbf{OLA}} & \multicolumn{ 2}{c}{\textbf{Rank}} & \multicolumn{ 2}{c}{\textbf{Knora-E}} & \multicolumn{ 2}{c}{\textbf{Knora-U}} & \multicolumn{ 2}{c}{\textbf{Meta-Des}} \\ 
 \hline
\textbf{Dataset} & \textbf{Bagging} & \textbf{PGDCS} & \textbf{Bagging} & \textbf{PGDCS} & \textbf{Bagging} & \textbf{PGDCS} & \textbf{Bagging} & \textbf{PGDCS} & \textbf{Bagging} & \textbf{PGDCS} & \textbf{Bagging} & \textbf{PGDCS} & \textbf{Bagging} & \textbf{PGDCS} \\ 
\hline
\textbf{01} & 86.9(2.1) & \textbf{88.1(2.3)*} & 82.9(2.7) & \textbf{84.0(2.8)} & 82.9(1.9) & \textbf{83.3(2.5)} & 81.4(2.7) & \textbf{81.5(2.8)*} & 82.2(3.1) & \textbf{82.3(2.8)} & 86.9(2.3) & \textbf{88.1(2.0)} & \textbf{86.2(2.5)} & 86.0(2.6) \\
\textbf{02} & \textbf{84.6(1.4)} & 83.5(2.0) & \textbf{95.7(0.9)} & \textbf{95.7(1.0)} & \textbf{97.4(0.6) }& \textbf{97.4(0.6)} & \textbf{96.9(0.8)} & \textbf{96.9(0.7)*} &\textbf{ 96.9(0.7)} & \textbf{96.9(0.6)} & \textbf{96.2(1.0)} & \textbf{96.2(0.9)} & 93.3(1.2) & \textbf{94.9(1.4)*} \\
\textbf{03} & \textbf{77.6(2.0)} & 76.7(1.9) & 70.4(5.3) & \textbf{72.9(5.3)} & 74.8(2.5) & \textbf{75.7(3.0)} & 69.8(3.7) & \textbf{70.3(3.6)} & \textbf{70.2(3.8)} & 70.1(3.3) & \textbf{78.7(1.6)} & 78.0(1.7) & 76.0(0.9) & \textbf{76.2(1.1)} \\
\textbf{04} & 81.1(24.3) & \textbf{84.7(17.7)} & 86.2(1.9) & \textbf{86.7(1.7)*} & 89.0(1.4) & \textbf{89.2(1.3)} & 88.9(1.6) & \textbf{89.3(1.3)} & 89.5(1.6) & \textbf{90.2(1.3)} & 89.0(2.1) & \textbf{89.7(1.2)*} & 88.8(1.1) & \textbf{89.6(1.2)*} \\
\textbf{05} & \textbf{76.8(2.8)} & 76.4(2.7) & \textbf{70.7(3.1)} & 69.8(3.2) & 72.4(2.6) & \textbf{73.1(3.1)} & \textbf{71.5(2.6)} & \textbf{71.5(3.3)} & \textbf{72.5(2.7)} & 72.3(2.5) & 76.9(2.7) & \textbf{77.0(2.8)} & 73.7(2.5) & \textbf{74.4(2.6)} \\
\textbf{06} & 69.4(1.7) & \textbf{70.5(1.4)} & 62.8(2.4) & \textbf{65.0(2.4)} & 67.7(1.9) & \textbf{68.7(1.5)*} & 67.4(2.1) & \textbf{68.1(1.6)} & 68.9(1.7) & \textbf{69.9(1.4)*} & 71.7(1.5) & \textbf{72.4(1.3)} & 70.3(1.4) & \textbf{70.8(1.9)} \\
\textbf{07} & 75.9(2.3) & \textbf{76.5(3.0)} &\textbf{ 69.1(2.2)} & \textbf{69.1(3.1)} & 71.2(2.3) & \textbf{71.3(2.6)} & \textbf{71.0(2.7)} & 70.1(2.5) & \textbf{72.4(2.3)} & 71.5(2.2) & 76.1(2.4) & \textbf{76.7(2.3)} & 72.9(1.7) & \textbf{73.8(1.8)} \\
\textbf{08} & 75.0(2.3) & \textbf{76.8(2.7)} & 69.0(7.8) & \textbf{72.8(3.7)} & \textbf{69.7(4.5)} & 68.4(4.9) & \textbf{65.7(5.2)*} & 64.7(6.2) & 65.6(6.1) & \textbf{66.4(5.6)} & 75.1(2.1) & \textbf{75.7(2.2)} & \textbf{74.0(1.5)} & 73.5(1.7) \\
\textbf{09} & 82.5(3.7) & \textbf{85.4(3.6)*} & 75.7(7.1) & \textbf{80.6(5.2)} & 77.5(4.5) & \textbf{78.4(5.7)} & \textbf{77.1(4.5)*} & 76.9(5.0) & \textbf{79.4(4.2)*} & 78.9(4.3) & 82.8(3.8) & \textbf{85.1(3.4)} & 82.9(4.0) & \textbf{83.1(3.9)} \\
\textbf{10} & 71.4(2.7) & \textbf{72.7(3.3)*} & 66.8(4.8) & \textbf{66.9(3.2)} & 69.0(3.5) & \textbf{69.1(3.3)} & \textbf{68.0(2.8)} & 67.8(3.4) & \textbf{68.9(2.3)} & 68.8(2.8) & 71.0(2.4) & \textbf{72.4(2.1)} & \textbf{70.7(1.6)} & 70.1(2.1) \\
\textbf{11} & 89.2(2.9) & \textbf{91.4(3.8)*} & 85.4(3.9) & \textbf{86.1(3.6)*} & 85.9(2.7) & \textbf{88.4(3.0)*} & 86.0(2.8) & \textbf{88.3(3.2)*} & 88.4(3.6) & \textbf{90.0(3.3)} & 89.2(2.8) & \textbf{91.5(3.8)*} & 88.8(3.0) & \textbf{89.8(3.2)*} \\
\textbf{12} & 83.5(3.7) & \textbf{84.2(3.7)*} & \textbf{77.6(6.8)*} & 75.1(5.2) & 80.6(3.7) & \textbf{81.4(4.7)} & 78.6(3.5) & \textbf{81.4(3.8)} & 79.9(3.7) & \textbf{82.5(4.4)*} & 83.0(3.9) & \textbf{84.1(3.7)*} & 83.0(5.5) & \textbf{84.3(5.6)} \\
\textbf{13} & 72.7(2.3) & \textbf{74.8(2.7)*} & 68.1(4.4) & \textbf{70.5(4.9)*} & \textbf{68.0(4.8)} & 67.9(3.8) & 64.1(5.8) & \textbf{66.6(4.3)} & 66.7(4.0) & \textbf{68.5(4.5)} & 73.5(2.1) & \textbf{75.9(2.6)*} & 71.1(2.8) & \textbf{72.1(3.2)} \\
\textbf{14} & 82.8(1.7) & \textbf{82.9(1.7)*} & 92.7(1.0) & \textbf{92.8(1.1)} & \textbf{96.7(0.7)} & 96.6(0.8) & \textbf{95.9(0.9)} & \textbf{95.9(0.8)*} & \textbf{96.0(0.7) }& \textbf{96.0(0.7)} & \textbf{95.1(1.1)} & 94.8(1.1) & \textbf{94.2(1.1)} & 94.1(1.0) \\
\textbf{15} & \textbf{68.7(4.2)} & 68.4(3.8) & 55.3(5.3) & \textbf{58.0(4.6)} & \textbf{68.2(3.9)} & 65.5(5.4) & \textbf{66.4(4.1)*} & 65.6(5.5) & \textbf{67.0(4.1)*} & 66.2(4.6) & \textbf{70.1(3.5)} & \textbf{70.1(5.2)} & 67.2(4.7) & \textbf{68.2(4.1)} \\
\textbf{16} & 82.9(2.1) & \textbf{83.8(2.5)} & 79.7(3.6) & \textbf{80.4(2.8)} & \textbf{80.8(2.6)} & \textbf{80.8(2.5)} & \textbf{77.4(2.4)} & 76.8(3.2) & 77.2(2.7) & \textbf{77.6(2.9)} & 82.9(1.9) & \textbf{83.3(2.3)} & \textbf{79.7(3.0)} & 79.6(3.3) \\
\textbf{17} & 78.9(3.2) & \textbf{83.4(2.6)*} & 72.8(4.3) & \textbf{73.8(4.7)} & 85.7(3.6) & \textbf{87.8(3.1)} & 85.5(3.8) & \textbf{87.6(3.7)} & 88.9(3.1) & \textbf{89.5(2.9)} & 81.8(3.5) & \textbf{84.6(2.7)*} & 91.5(4.5) & \textbf{94.1(2.9)*} \\
\textbf{18} & 75.6(1.1) & \textbf{76.1(1.4)} & 75.7(2.4) & \textbf{76.2(1.9)} & 82.8(0.6) & \textbf{83.0(0.9)} & 83.0(0.8) & \textbf{83.3(0.9)} &\textbf{ 83.5(0.9)} & \textbf{83.5(0.8)} & 81.8(1.0) & \textbf{81.9(0.9)} & \textbf{82.5(0.9)} & 82.4(0.9) \\
\textbf{19} & 56.5(4.6) & \textbf{56.7(4.4)*} & 73.4(3.0) & \textbf{74.0(2.7)*} & \textbf{88.1(1.7)} & 88.0(2.2) & \textbf{90.1(1.7)} & 89.8(2.3) & \textbf{90.5(1.6)} & 89.5(2.2) & \textbf{87.2(2.8)} & 85.8(2.3) & \textbf{86.8(2.5)} & \textbf{86.8(2.6)}  \\
\textbf{20} & \textbf{91.0(1.2)} & \textbf{91.0(1.4)} & 88.7(1.2) & \textbf{88.8(1.2)} & \textbf{92.8(1.4)} & 92.7(1.2) & \textbf{93.0(1.1) }& \textbf{93.0(1.2)} & \textbf{94.2(1.0)} & \textbf{94.2(1.0)} & 92.3(1.0) & \textbf{92.4(1.4)} & 94.0(1.1) & \textbf{94.3(0.9)} \\
\textbf{21} & 77.7(3.4) & \textbf{81.7(6.5)*} & 66.9(6.8) & \textbf{73.3(7.4)*} & 75.1(6.2) & \textbf{78.6(6.9)*} & 74.8(5.5) & \textbf{79.6(6.1)*} & 78.5(3.5) & \textbf{82.2(4.5)*} & 78.1(3.0) & \textbf{82.1(6.4)*} & 79.7(3.7) & \textbf{82.7(6.0)*} \\
\textbf{22} & 96.6(1.2) & \textbf{97.3(0.9)*} & 95.0(1.9) & \textbf{95.5(1.6)*} & 95.7(1.7) & \textbf{96.0(1.5)} & 95.1(1.7) & \textbf{95.7(1.8)*} & 95.9(1.8) & \textbf{96.3(1.5)} & 97.0(1.0) & \textbf{97.4(0.9)*} & 96.4(0.9) & \textbf{96.7(0.8)} \\
\textbf{23} & 75.2(2.0) & \textbf{77.5(2.4)*} & 67.1(2.5) & \textbf{69.1(3.0)} & 73.1(2.0) & \textbf{74.0(2.7)} & 73.0(2.0) & \textbf{74.1(3.1)} & \textbf{75.8(2.2)} & \textbf{75.8(2.6)*} & 76.0(2.1) & \textbf{77.7(2.1)} & 75.9(2.3) & \textbf{76.6(2.4)} \\
\textbf{24} & 87.1(3.4) & \textbf{87.9(3.6)} & 81.7(4.6) & \textbf{82.7(3.4)} & \textbf{83.8(4.3)} & \textbf{83.8(3.9)} & \textbf{84.3(5.0)} & 83.6(3.6) & 84.8(4.3) & \textbf{85.1(3.5)} & 86.5(4.0) & \textbf{87.1(3.5)} & 86.7(3.8) & \textbf{87.3(3.0)} \\
\textbf{25} & 97.1(1.1) & \textbf{98.2(1.0)*} & 95.0(1.4) & \textbf{95.3(1.8)} & 96.0(1.3) & \textbf{96.6(1.3)*} & 95.9(1.3) & \textbf{96.7(1.3)} & 97.1(0.9) & \textbf{97.7(1.2)} & 97.1(1.2) & \textbf{98.2(1.0)*} & 96.9(1.3) & \textbf{97.5(1.6)} \\
\textbf{26} & 86.5(0.8) & \textbf{86.8(0.9)*} & \textbf{79.2(2.1)} & 78.9(1.8) & 82.4(1.2) & \textbf{82.9(0.9)} & 82.0(0.9) & \textbf{82.3(0.9)*} & 83.4(0.8) & \textbf{83.8(1.0)} & 86.4(0.8) & \textbf{86.8(1.0)} & \textbf{85.9(0.8)} & \textbf{85.9(0.9)} \\
\textbf{27} & 82.5(3.8) & \textbf{84.9(3.6)*} & 75.4(5.0) & \textbf{77.3(4.4)} & 79.7(4.5) & \textbf{80.3(5.2)} & 79.9(4.7) & \textbf{80.1(4.9)} & \textbf{82.1(4.6)} & 81.9(4.7) & 83.2(3.6) & \textbf{85.5(3.7)} & 82.7(3.4) & \textbf{84.4(4.1)} \\
\textbf{28} & 97.5(2.3) & \textbf{98.0(2.3)} & 94.2(4.3) & \textbf{96.1(2.4)} & 95.0(2.2) & \textbf{95.6(2.7)} & 95.0(2.2) & \textbf{95.6(2.7)} & \textbf{97.5(2.0)} & 97.3(2.4) & 97.5(2.1) & \textbf{98.0(2.3)} & 98.0(2.0) & \textbf{98.2(2.2)} \\ \hline
\textbf{Aver.} & \multicolumn{1}{r}{80.83} & \textbf{82.01} & \multicolumn{1}{r}{77.61} & \textbf{78.84} & \multicolumn{1}{r}{81.5} & \textbf{81.95} & \multicolumn{1}{r}{80.63} & \textbf{81.18} & \multicolumn{1}{r}{81.93} & \textbf{82.32} & \multicolumn{1}{r}{83.68} & \textbf{84.59} & \multicolumn{1}{r}{83.21} & \textbf{83.84}  \\ 
\textbf{Diff.} &  & -1.18 &  & -1.23 &  & -0.45 &  & -0.55 &  & -0.39 &  & -0.91 &  & -0.63 \\ \hline
\end{tabular}
}
\end{table*}